\documentclass[letterpaper, 10 pt, conference]{ieeeconf}  

\IEEEoverridecommandlockouts                              

\usepackage[dvipdfmx]{graphicx} 
\usepackage{epsfig} 
\usepackage{times} 
\usepackage{amsmath} 
\usepackage{amssymb}  
\usepackage{mathtools}
\usepackage{cite}
\usepackage{color}

\title{\LARGE \bf Tactile Perception based on Injected Vibration in Soft Sensor}
\author{Naoto Komeno and Takamitsu Matsubara
\thanks{*This work was supported by JSPS KAKENHI Grant Number 19H01124.}
\thanks{All authors are with the Division of Information Science, Graduate School of Science and Technology, Nara Institute of Science and Technology, Japan:
        {\tt\small komeno.naoto.km6@is.naist.jp, takam-m@is.naist.jp}}%
}

\begin{document}

\maketitle
\thispagestyle{empty}
\pagestyle{empty}

\begin{abstract}
Tactile perception using vibration sensation helps robots recognize their environment's physical properties and perform complex tasks. 
A sliding motion is applied to target objects to generate tactile vibration data. 
However, situations exist where such a sliding motion is infeasible due to geometrical constraints in the environment or an object's fragility which cannot resist friction forces. 
This paper explores a novel approach to achieve vibration-based tactile perception without a sliding motion. 
To this end, our key idea is injecting a mechanical vibration into a soft tactile sensor system and measuring the propagated vibration inside it by a sensor. 
Soft tactile sensors are deformed by the contact state, and the touched objects' shape or texture should change the characteristics of the vibration propagation. 
Therefore, the propagated-vibration data are expected to contain useful information for recognizing touched environments. 
We developed a prototype system for a proof-of-concept: a mechanical vibration is applied to a biomimetic (soft and vibration-based) tactile sensor from a small, mounted piezoelectric actuator.  
As a verification experiment, we performed two classification tasks for sandpaper's grit size and a slit's gap widths using our approach and compared their accuracies with that of using sliding motions. 
Our approach resulted in 70$\mathrm{\%}$ accuracy for the grit size classification and 99$\mathrm{\%}$ accuracy for the gap width classification. These results are comparable to or better than the comparison methods with a sliding motion. 
\end{abstract}

\section{INTRODUCTION}
With their sense of touch, humans can perceive physical interactions with the outside world, including contact with objects, their shape, texture, hardness, and temperature.
Based on this tactile information, we can recognize the physical properties of the environment and perform complex tasks.
Humans use six types of motion strategies to extract information about the properties of objects during tactile exploration \cite{HP_tutorial}.
In robot control, there is a growing body of research on exploiting tactile information to perform tasks \cite{Cui2017,PFN} as well as research on tactile search \cite{AHPR,Fishel2012,Tanaka2014-manifold,Tanaka2014-optimal}.

Vibration is one physical interaction that produces such tactile sensations.
Vibrations, which are generated by touching an object, provide essential tactile elements such as contact detection, texture, hardness, and slip.
The engineering approach to vibration is relatively easy because many signal processing techniques can be applied to it. 
It is also less problematic in terms of size and integration density for developing tactile sensors to detect vibrations than distributed features \cite{Shimonomura2019}.
From this perspective, the sense of vibration has a high degree of engineering applicability and is being studied in combination with robot control \cite{sewers,Su2015,Zou2017}.

\begin{figure}[tbp]
  \centering
  \includegraphics[width=\hsize]{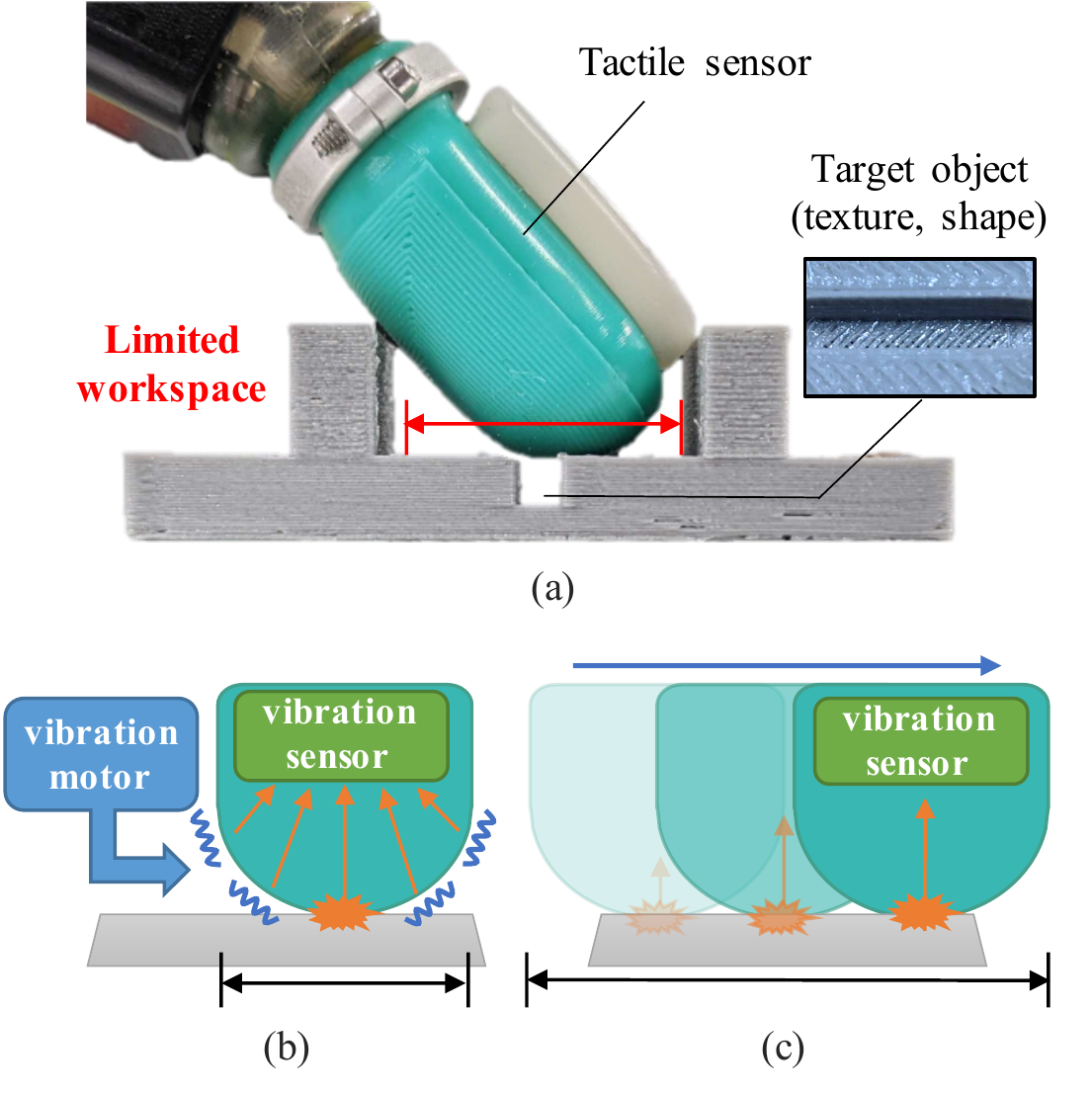}
  \caption{Concept of our approach: (a) Robot's motion is restricted in a narrow space and no vibration occurs. In such a situation, it cannot detect any tactile information. (b) Our method solves this difficulty by injecting vibration. (c) Conventional sliding motion must be comprehensive. }
  \label{fig: senzai}
\end{figure}

On the other hand, to obtain a sense of vibration, sliding motions are essential for tactile search behavior patterns \cite{HP_tutorial}.
However, strong geometrical and dynamic constraints occur in the robot when it makes contact with an object.
For example, in a narrow space, such as inside a curved pipe, the robot's joints interfere with the surrounding objects, which complicates achieving a high degree of freedom of motion control \cite{collision}.
In surgical procedures, the robot requires kinetic control to avoid damaging the surrounding biological tissues which cannot resist friction forces \cite{narrow}. 
These restrictions of sliding motion for generating tactile vibration data motivated us to explore another approach 
that allows vibration-based tactile perception without sliding motions. 

This paper investigates a novel approach for vibration-based tactile perception without a sliding motion (Fig~\ref{fig: senzai}).
We aim to achieve vibration-based tactile perception by applying mechanical vibrations to a soft tactile sensor system and measuring the vibrations propagated by internal sensors.
Since the soft tactile sensor deforms according to the contact's state, the vibration's propagation characteristics are different, depending on the shape and texture of the touched object.
Therefore, we expect that the propagated-vibration data will contain useful information for recognizing touched environments.
We developed a prototype system that applies mechanical vibrations to a biomimetic tactile sensor equipped with a small piezoelectric actuator for a proof-of-concept.
As a verification experiment, we performed two classification tasks for sandpaper's grit size and a slit's gap widths using our approach and compared their accuracies with that of using sliding motions. 
Our approach increased the accuracies for both tasks to a level that was comparable to or better than the comparison methods a sliding motion. 

The remainder of this paper is organized as follows. 
Section II summarizes related works. 
Section III describes our proposed approach and a prototype system for subsequent experimental verification. 
Section IV presents our conducted experiments and obtained results. Sections V and VI discuss and conclude this paper.

\section{RELATED WORKS}
As an example of research on classification tasks using BioTac, Delhaye et al. investigated how the texture classification rate changes with tracing speed in various materials \cite{Delhaye2016}.
Also, Fishel et al. proposed an active classification method for tracing textures using a Bayesian exploration framework \cite{Fishel2012}.

Besides, several tactile sensing methods utilize the properties of vibration propagation in a structure.
Kuang et al. used an eccentric motor to vibrate a cantilevered beam to perform force sensing with an accelerometer \cite{Kuang2020}. In their method, the acceleration caused by the vibration is given as an offset baseline value to convert the stress change into an acceleration change to estimate the force vector.
This method allows more movement of the sensor base away from the contact surface than conventional force sensors, reducing the risk of damage and simplifying the end-effector.
However, although accelerometers have a high sampling frequency, they are noisy and have low resolution.
Shinoda et al. developed an Acoustic Resonance Tensor Cell (ARTC) as a tactile sensor using ultrasonic echoing properties \cite{Shinoda1997,Ando2001}.
This sensor has a soft surface and can detect deformation and slips by changing the inner cavity's resonance frequency.
Due to the high performance of ultrasonic measurements, such small stress changes as incipient slips can be captured.
A method to identify friction coefficients from instantaneous contact is also being studied using this sensor \cite{Shinoda2000}. Inspired by these studies, we use the characteristics of vibration propagation instead of sliding motions for tactile sensing. However, our approach differs because it targets such tactile information as texture and shape instead of stress. Moreover, our proposed method can be implemented quickly if the tactile sensors meet the required specifications.

Our study is also different from tactile sensor systems that use stochastic resonance, which is a phenomenon where a weak sub-threshold signal is amplified in nonlinear dynamics with noise \cite{Moss2004}.
This phenomenon has been widely observed in nature. 
Stochastic resonance occurs in human tactile sensation due to such dynamics as the threshold system of mechanoreceptors.
Kurita et al. attached a small piezoelectric actuator to a glove to improve the sensorimotor ability of human fingertips by applying vibration \cite{Kurita2013}.
Ohka et al. improved the signal-to-noise ratio by giving strong nonlinearity to the signal processing circuit of a force sensor \cite{Ohka2009}.
Although these studies closely resemble the present study in terms of vibration in tactile sensing, their underlying mechanism is quite different.
The purpose of this research is tactile sensing without sliding motion, and there are no sub-threshold signals to be amplified.
We use white noise not to amplify other signals, but to focus on the propagation characteristics themselves for analysis.
This is a crucial difference from stochastic resonance.

\begin{figure}[tbp]
  \centering
  \includegraphics[width=0.9\hsize]{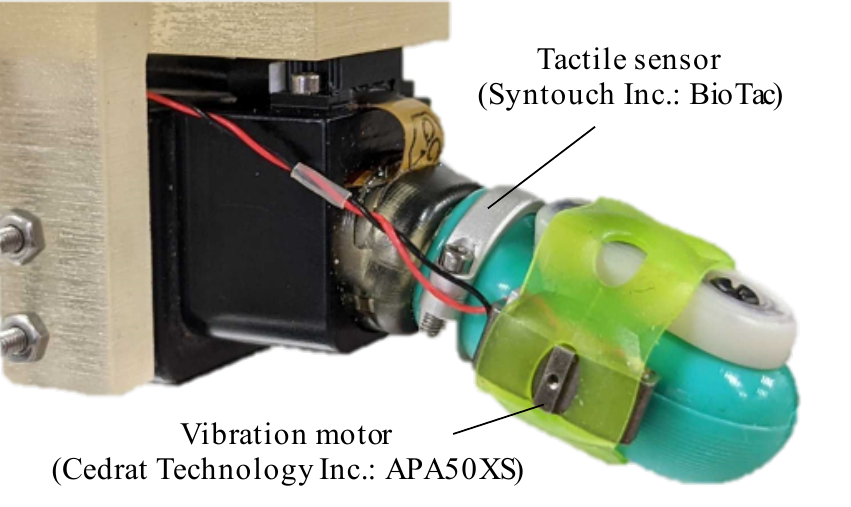}
  \caption{Prototype system of injected vibration-based tactile perception: We attached a small piezoelectric actuator to the side of biomimetic tactile sensor by finger cot.}
  \label{fig: realsystem}
\end{figure}

\section{PROPOSED SYSTEM}

    \subsection{Mechanism}
    We propose a tactile sensing system using a soft tactile sensor and a vibration motor.
    The mechanism of the proposed method is shown in Fig. \ref{fig: senzai}(b).
    We apply vibration from an external vibration motor to a tactile sensor covered with a soft material.
    Vibrations propagate through the soft structure and are measured by internal vibration sensors.
    The vibrations vary in propagation characteristics depending on the soft mediating structure state and resemble a change of time-series waves or a frequency spectrum.
    If the tactile sensor does not touch anything, the state of the soft structure cannot be affected by anything.
    By contrast, when it does contact an object, the contact area and deformation shape change depending on the object's roughness, stickiness, shape, and hardness; the vibration propagation characteristics also change.
    The propagation properties, which contain tactile information, are converted to tactile information by a machine-learning classifier.
    
    To achieve the proposed method, the soft structure of the tactile sensor requires low viscosity and sufficient volume so that the applied vibration propagates well.
    For detecting small changes in signal components, a highly sensitive vibration sensing capability may also be required.
    Using white noise signals is also beneficial for system identification, e.g., Gaussian noise and M-sequence signals for vibration.
    This is because data-driven tactile information is collected without considering the sensor or object model by applying a signal that excites all the system modes.
    Actuators with sufficient power over a wide frequency range are required to output these signals.

\begin{figure}[tbp]
  \centering
  \includegraphics[width=\hsize]{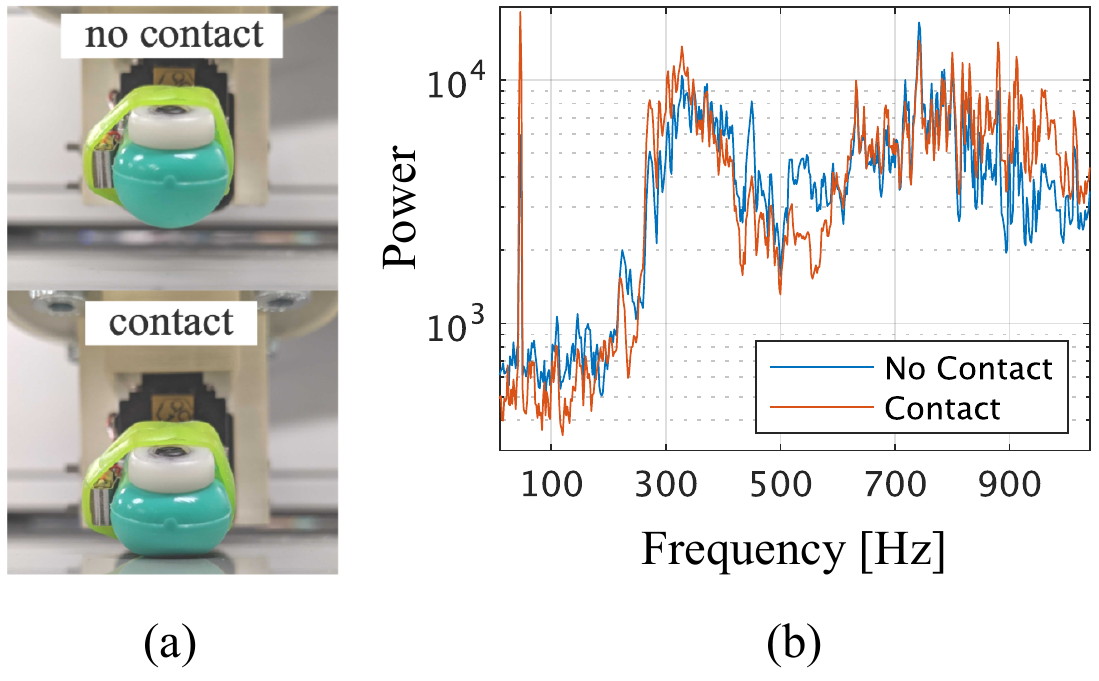}
  \caption{Result of preliminary experiment: (a) scenes of contact and no contact and (b) comparison of frequency spectrum with and without contacting the table using our prototype. Frequency spectrum varies with contact states with injection of Gaussian noise (I=0 dB).}
  \label{fig: preliminary}
\end{figure}

    \subsection{Prototype}
    We built an actual sensor system. 
    We selected a tactile sensor and a vibration motor that satisfy our proposed method's requirements.
    The system we built is shown in Fig. \ref{fig: realsystem}.

        \subsubsection{Tactile sensor}
        We used a biomimetic tactile sensor (Syntouch, BioTac) with a soft structure.
        BioTac, which is a tactile sensor that imitates a human finger, is a multimodal sensor capable of measuring pressure and its distribution, temperature, and heat flux \cite{bt_fab,bt_human}.
        Its surface is covered with artificial skin imprinted with fingerprints and filled with a conductive fluid.
        This gives BioTac the softness of a human finger and allows it to make gentle contact with an object.
        The pressure and vibration generated at the contact surface can be transmitted as pressure through the fluid and sensed by the internal hydroacoustic pressure sensor \cite{bt_fab,bt_microvib}.
        This structure detects the micro-vibrations generated at the sensor surface with pressure resolution up to 0.37-Pa and a 2.2-kHz sampling frequency.
        These features satisfy two requirements of the proposed system: soft structure and vibration detection.

        \subsubsection{Vibration motor}
        We adopted a piezoelectric actuator, which is a vibration motor capable of outputting a white noise signal.
        Piezoelectric actuators are frequently used in stochastic resonance experiments because they have a sweeping drive frequency.
        Kurita et al. developed a glove that enhances the human fingertips' tactile sensitivity by applying vibration with a small piezoelectric actuator \cite{Kurita2013}.
        Since BioTac's size is also close to that of a human finger, we also applied vibrations with a small piezoelectric actuator (Cedrat technology, APA50XS) that
        is mounted on the BioTac's side (Fig. \ref{fig: realsystem}).

        The vibrations are generated from the standard normal distribution $\mathcal{N}(0,1)$ of Gaussian noise $\varepsilon$ at a noise intensity of I dB:
        \begin{equation}
          \varepsilon \sim 10^{I/20}\mathcal{N}(0,1).
          \label{eq: noise}
        \end{equation}
        We applied a low-pass filter with a cut-off frequency of 1100 Hz to match the sensor's sampling frequency.
        In the experiments, we used noise intensity I as the experiment's parameter.

    \subsection{Preliminary Experiment}
    We conducted preliminary experiments with the proposed method and
   verified whether the signal components changed when the sensor contacts the table.
    We injected a white noise of the maximum motor power rating (I=0 dB) and got vibration data for 50 seconds.
    We calculated the frequency spectrum and observed the differences.
    The preliminary experimental results are shown in Fig. \ref{fig: preliminary}.
   Since the frequency spectrum changed depending on the contact,
    the contact state can be estimated by examining the frequency spectrum of the applied vibration.

\section{EXPERIMENT}

    \subsection{Tasks}
    We verified whether the proposed method can be used for tactile sensing.
    In our experiments, we performed two classification tasks on sandpaper's grit sizes and a slit's gap widths.
    Since sandpaper is widely used in psychological experiments and tactile sensor performance experiments \cite{Kurita2013,sandpaper}, we used it as one example of texture classification.
    We prepared slits to imitate the gaps and steps between parts that occur in the assembly process of vehicles \cite{Lepora2016} and investigated whether our method can detect local shapes.
    
    The contact with the sample is made at the fixed coordinates set in beforehand, and after the data acquisition, it is confirmed that the hydrostatic pressure $\mathrm{P_{DC}}$ inside the BioTac does not change significantly due to the experimental conditions.
    
        \subsubsection{Grit size classification}
        We classified five different sandpapers with the following JIS grit sizes: J40, J80, J120, J240, and J400.
        The grit size indicates the number of particles within a 10-mm square.
        The sandpaper is attached in concentric circles to a 170-mm-diameter circular table manufactured by a 3D printer.
        The table is fixed to a servo motor (Robotis, XM430-W350-T).
        We can change the class at fixed coordinates by rotating it.
        
        \subsubsection{Gap width classification}
        We classified five different slit gaps of 0.5- to 4.5-mm widths in 1.0-mm equal steps.
        Each gap was engraved in concentric circles on a 170-mm-diameter circular table manufactured by a 3D printer. 
        This table can also be rotated.

\begin{figure}[tbp]
  \centering
  \includegraphics[width=0.9\hsize]{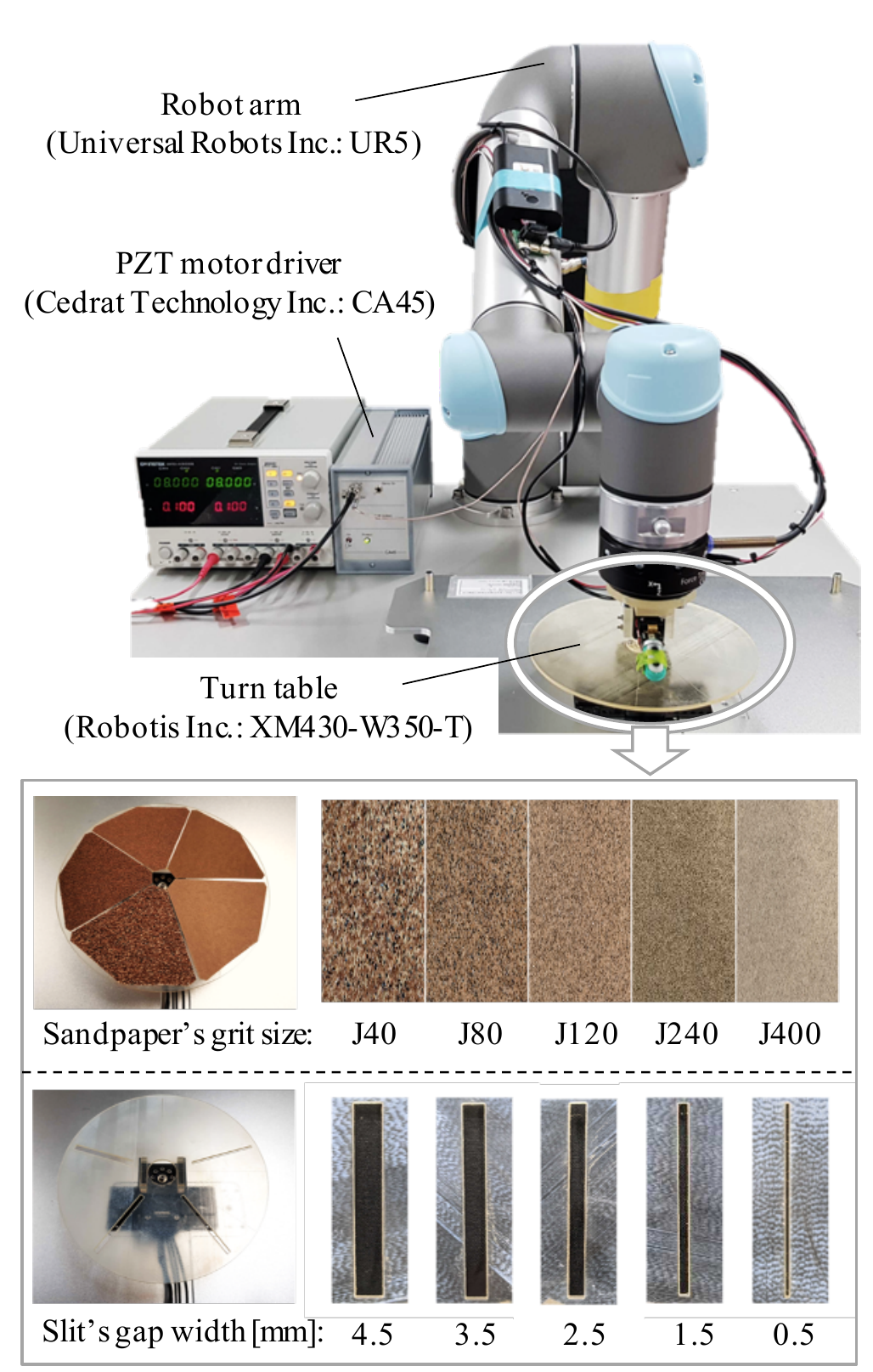}
  \caption{ Experiment setup: Circular tables were manufactured by a 3D printer. Sandpapers and slits were attached to them. Robotic arm with proposed system touches them to collect corresponding tactile data.}
  \label{fig: robot}
\end{figure}

    \subsection{Data Collection}
    
        \subsubsection{Proposed method}
        Figure \ref{fig: robot} shows the experimental environment.
        A finger cot (axis, colored finger cot M) is attached to the tactile sensor and mounted on a robot arm (Universal Robots, UR5).
        The finger cot has two roles: to protect the sensor and to be attached to the piezoelectric actuator.
        A vibration motor shakes the tactile sensor while it touches a target on the table.
        The robot arm stops in a fixed position and touches a target.
        Each trial takes 0.5 seconds, for a total of 100 trials.
        The vibration level is set based on noise intensity I in Eq. (\ref{eq: noise}), and the experiment is divided into seven levels.
        The relationship between the vibration level and the noise intensity is shown in Table \ref{tbl: noise}.
        We randomly set the order of the contact classes and applied the vibration levels.
        We also changed the class by rotating the table to maintain constant contact coordinates for the robot arm to prevent an arm pose (that derives the vibration information) from contaminating the sensor.
        During the contact, the power of the robot should be left active to shorten the experiment time.
        
\begin{table}[tb]
 \centering
 \caption{Details of vibration level}
 \label{tbl: noise}
 \begin{tabular}{|c||c|}
  \hline
	Vibration level	&I [dB]\\\hline\hline
	0	&-999\\\hline
    1	&-20\\\hline
    2	&-16\\\hline
    3	&-12\\\hline
    4	&-8\\\hline
    5	&-4\\\hline
	6	&0\\
  \hline
 \end{tabular}
\end{table}
        
        \subsubsection{Comparison}
        We performed another classification experiment for comparison. 
        Instead of applying vibration, a sliding motion was performed at 10 mm/s to obtain the tactile information.
        To avoid damaging the sensor system by friction, the finger cot was replaced every 25 samples.
        The sliding motion against the slit was performed longitudinally to keep the contact state the same during the motion.
        If in the vertical direction, there is less chance for the sensor to contact the slit, and the step information is incompatible with the frequency spectrum.

\begin{figure*}[tbp]
  \centering
  \includegraphics[width=0.9\hsize]{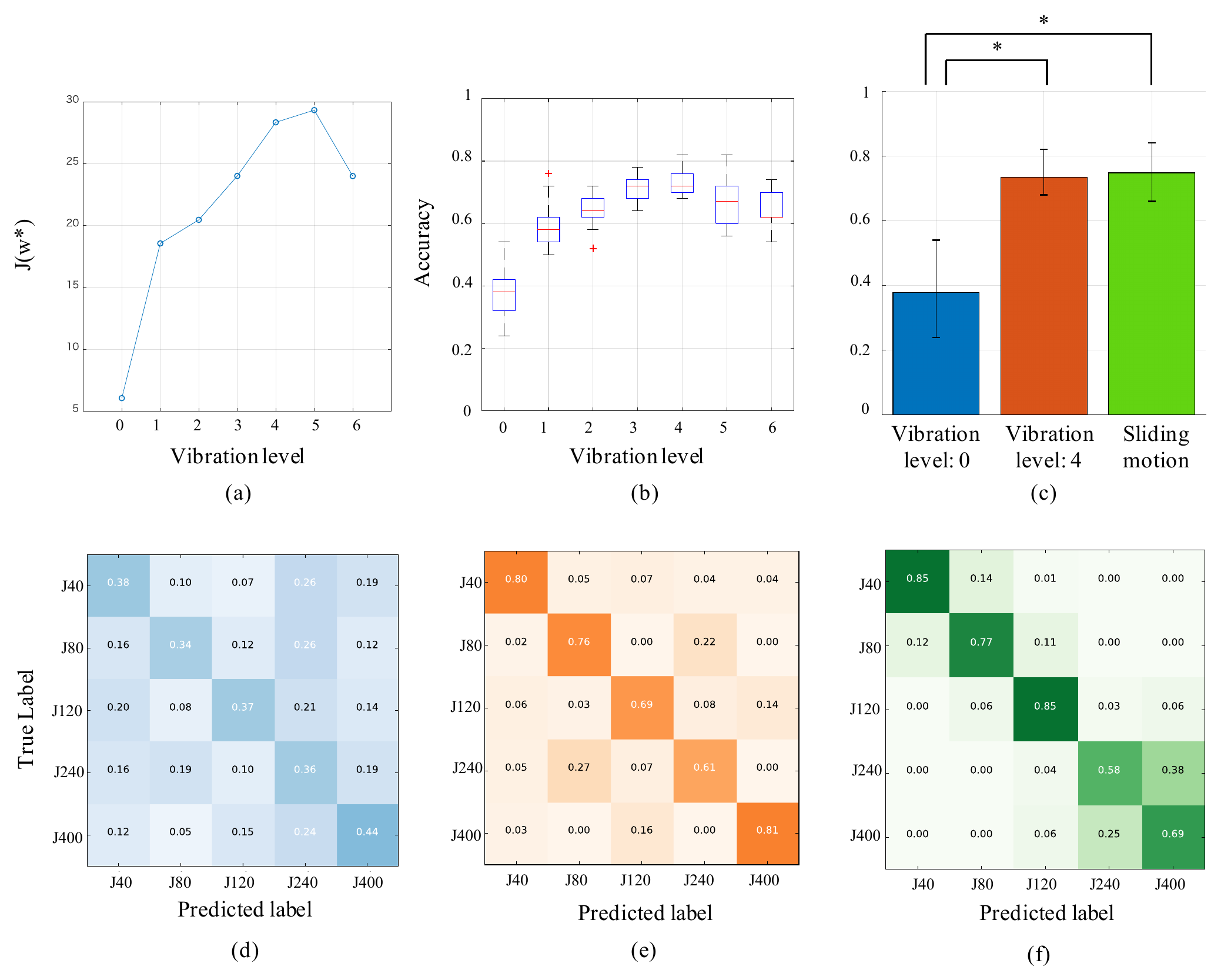}
  \caption{Results of grit size classification: (a) maximum class separation in each vibration level, (b) classification rate at each vibration level, (c) classification comparison at vibration levels: 0, 4, and sliding motion ($\ast$ means p $<$ 0.01), (d) confusion matrix of vibration level: 0, (e) confusion matrix of vibration level: 4, (f) confusion matrix of sliding motion.}
  \label{fig: result_sandpaper}
\end{figure*}

\begin{figure}[tbp]
  \centering
  \includegraphics[width=\hsize]{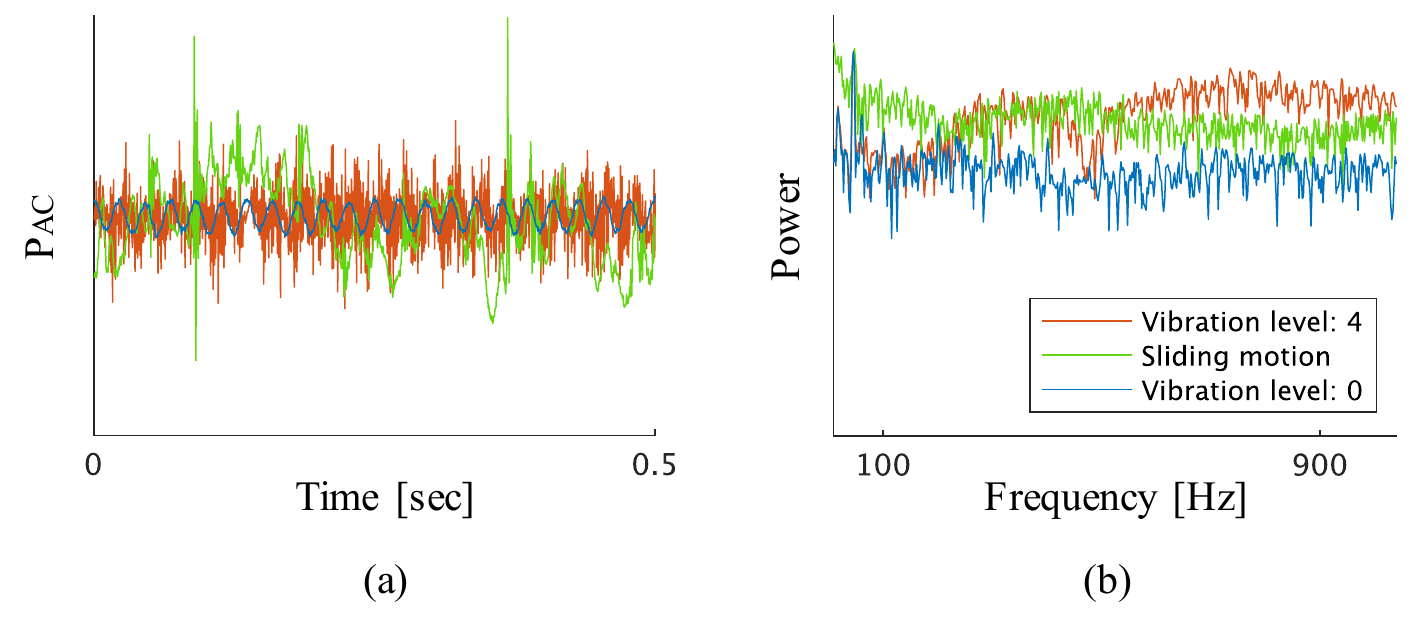}
  \caption{Output data from sensor in grit J40: (a) micro-vibration waveform and (b) micro-vibration frequency spectrum }
  \label{fig: wave_sandpaper}
\end{figure}

\begin{figure}[tbp]
  \centering
  \includegraphics[width=\hsize]{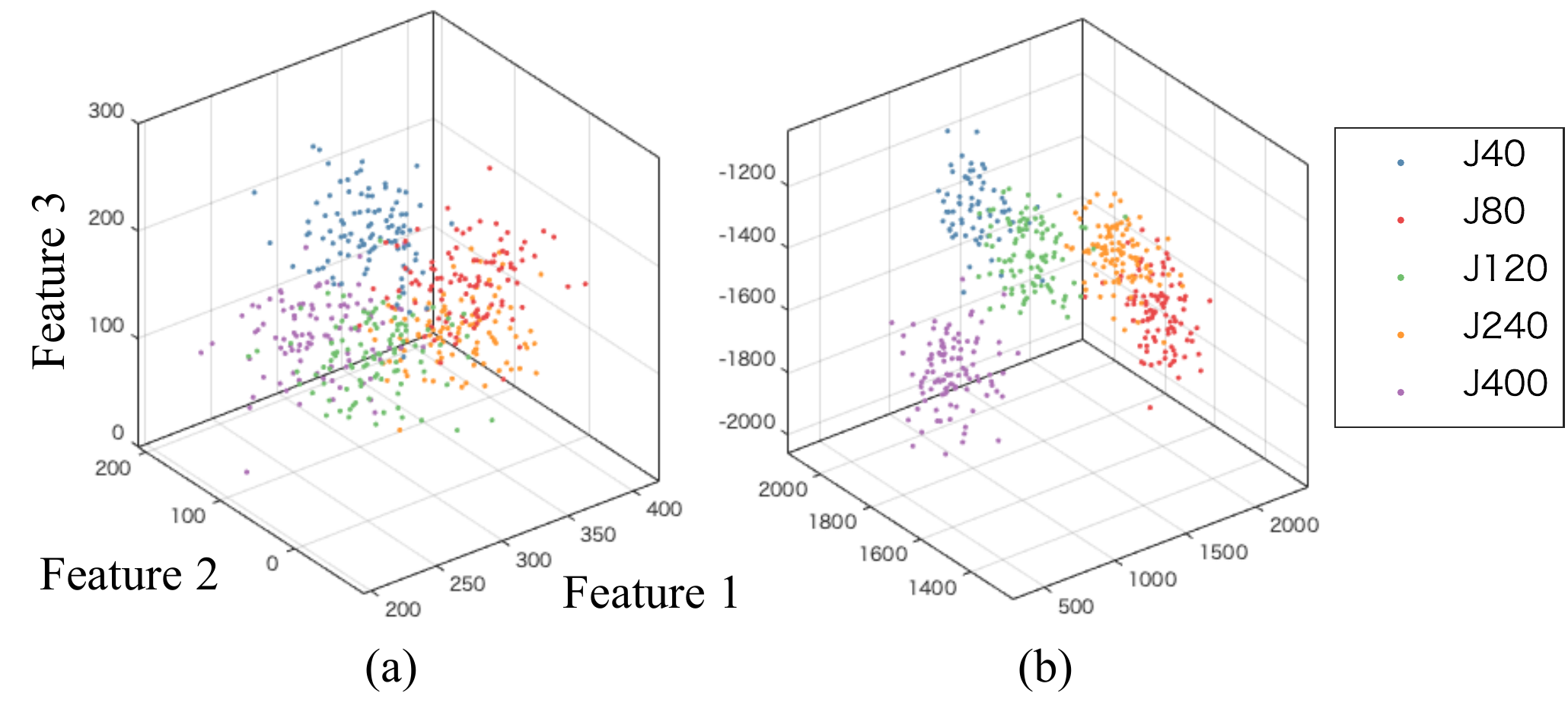}
  \caption{Feature space in grit size classification obtained by Fisher's linear discriminant in our method. The three largest eigenvalues determine features 1, 2, and 3. (a) Vibration level: 0 and (b) vibration level: 4}
  \label{fig: fspace_sandpaper}
\end{figure}

\begin{figure*}[tbp]
  \centering
  \includegraphics[width=0.92\hsize]{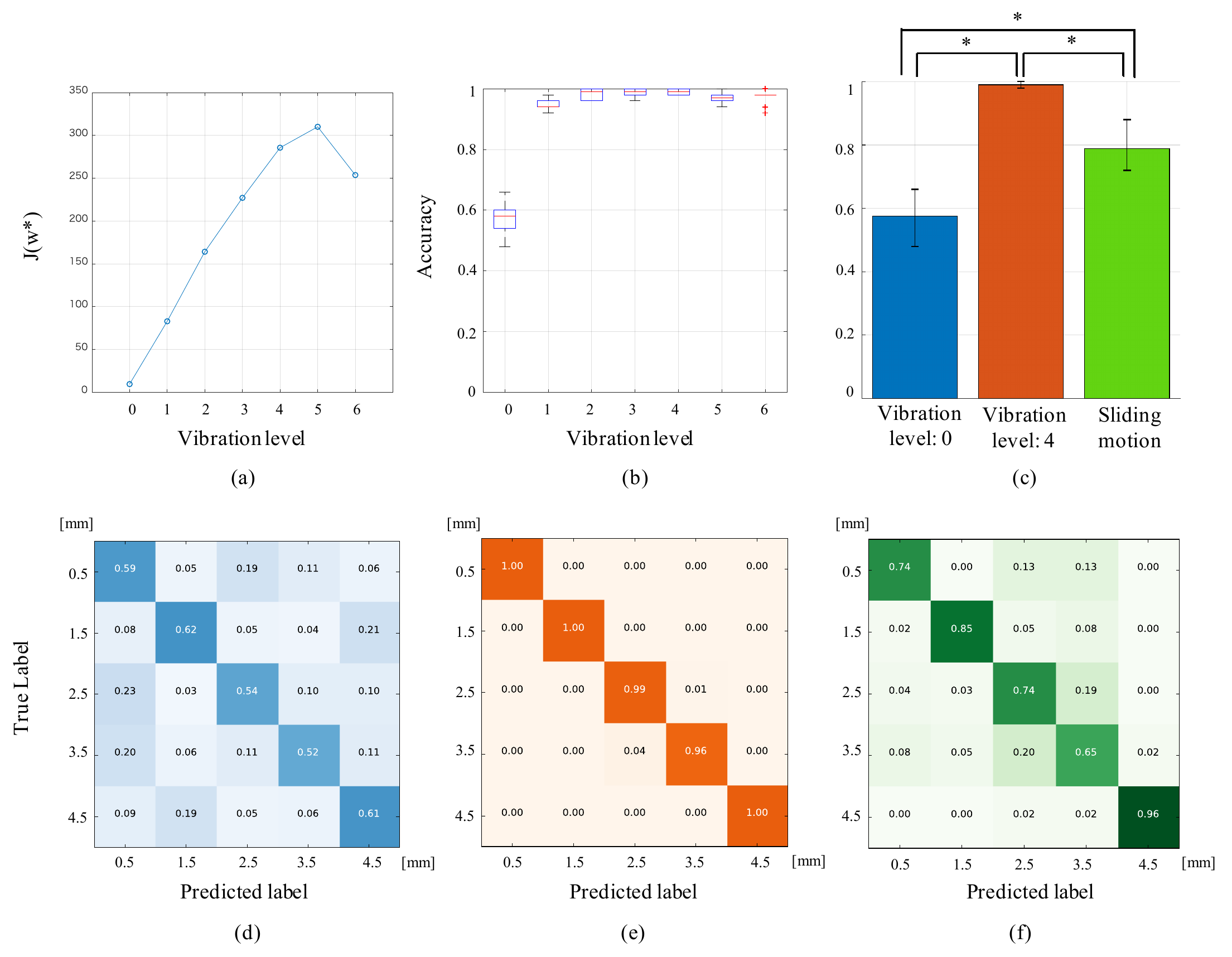}
  \caption{Results of gap width classification: (a) maximum class separation at each vibration level, (b) classification rate at each vibration level, (c) classification comparison at vibration levels: 0, 4, and sliding motion ($\ast$ means p $<$ 0.01), (d) confusion matrix of vibration level: 0, (e) confusion matrix of vibration level: 4, and (f) confusion matrix of sliding motion.}
  \label{fig: result_slit}
\end{figure*}

\begin{figure}[tbp]
  \centering
  \includegraphics[width=\hsize]{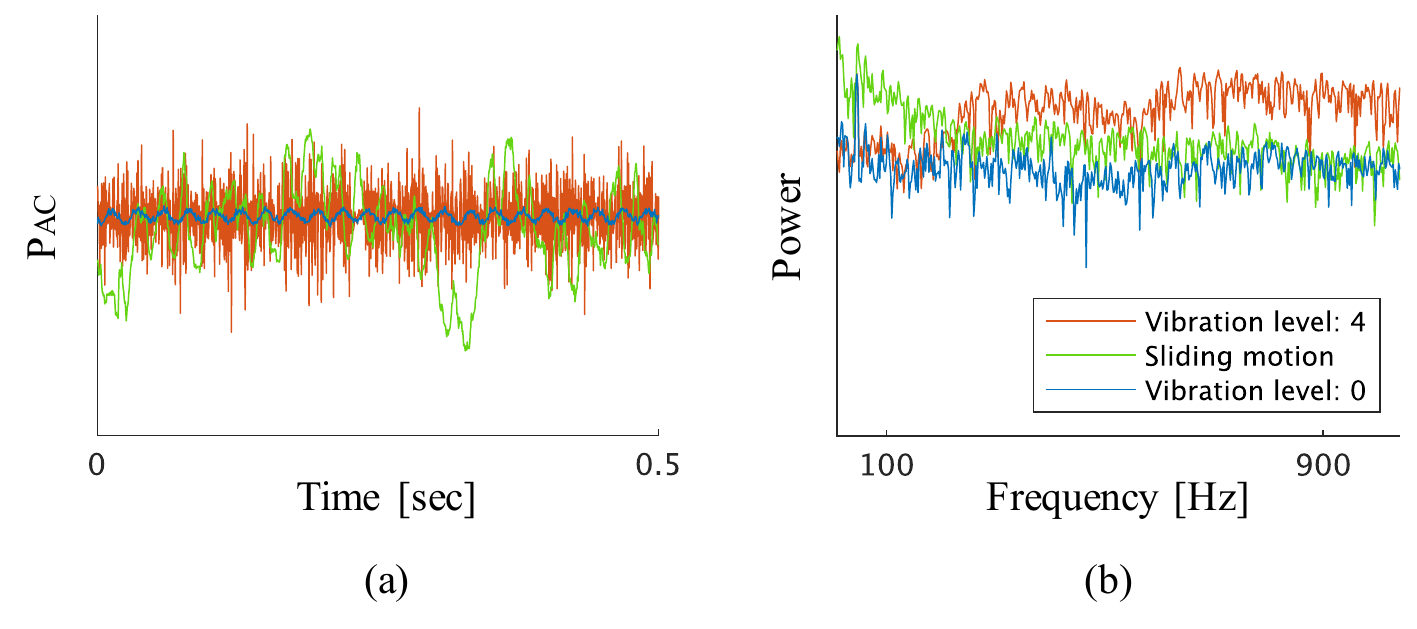}
  \caption{Output data from sensor at gap width 0.5-mm gap widths: (a) micro-vibration waveform and (b) micro-vibration frequency spectrum }
  \label{fig: wave_slit}
\end{figure} 

\begin{figure}[tbp]
  \centering
  \includegraphics[width=\hsize]{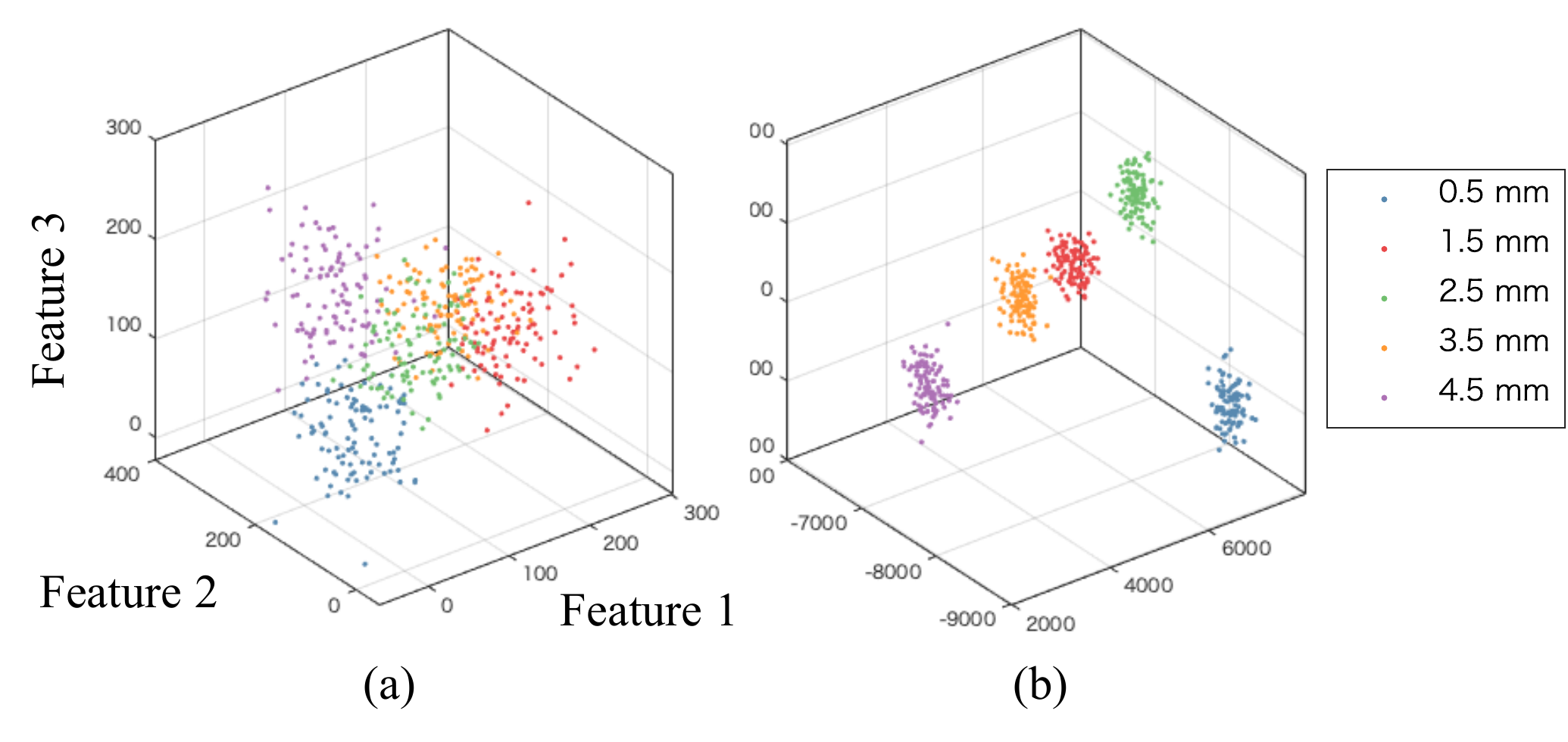}
  \caption{Feature space in gap width classification obtained by Fisher's linear discriminant in our method. The three largest eigenvalues determine features 1, 2, and 3. (a) Vibration level: 0 and (b) vibration level: 4}
  \label{fig: fspace_slit}
\end{figure}

    \subsection{Data Analysis}
    We extracted tactile information from the changes in the signal components of the applied vibrations due to the propagation.
    As a feature, we used the frequency spectrum of micro-vibration $\mathrm{P_{AC}}$ obtained from BioTac \cite{Delhaye2016,Kappassov2015}.
    The frequency spectrum is calculated by superimposing Hanning window on data with the length of 1100 per sample measured for 0.5 seconds at a sampling frequency of 2.2 kHz.
    We used the BioTac frequency response band: the 10-1040 Hz frequency band and obtained 515-dimensional frequency-spectrum vectors as features.
    We used the features to calculate the following maximum class separation and Support Vector Machine (SVM) classification rates for respectively evaluating the classification task in the linear and nonlinear space.
    We evaluated each of the vibration levels and compared them.
    
        \subsubsection{Maximum class separation}
        In Fisher's linear discriminant, we projected the data with $D$-dimensional features into a $D'$-dimensional feature space to maximize the separation of the classes \cite{PRML}. 
        For data with $N$ samples and $K$ classes, in the $k$th class of $D$-dimensional data $x_n$ with $N_k$ samples, we describe evaluation function $\mathrm{J(W)}$ of the Fisher's linear discriminant in the multiclass as follows:
        \begin{equation}
          \mathrm{J(W)} = \mathrm{Tr\{ (W S_W W^T)^{-1}(W S_B W^T) \}},
          \label{eq: J}
        \end{equation}
        where $\mathrm{W}$ is the projection matrix and $\mathrm{S_W, S_B}$ is the within-class and between-class covariance matrices.
        Eq. (\ref{eq: J}) expresses the degree of class separation in the projection space from the ratio of the between-class and within-class variances.
        Therefore, by maximizing evaluation function $\mathrm{J(W)}$, we can obtain projection matrix $\mathrm{W}^*$ that maximizes the class separation.
        Such a projection matrix $\mathrm{W}^*$ can be determined by the eigenvector of $\mathrm{S_W^{-1}S_B}$, which corresponds to the largest eigenvalue of $D'$ number of eigenvalues.
        
        In this study, maximum class separation $\mathrm{J(W^*)}$, obtained by maximizing Eq. (\ref{eq: J}), is evaluated as the entire dataset's classification sensitivity.
        The dimension of the projected feature space is $D'=3$, since the within-class covariance matrix's rank is high: $K-1$.
        This space can visualize how the obtained features are separable.
        
        \subsubsection{Classification rate}
        We classified grit sizes and gap widths by SVM with radial basis function kernel using the C-Support Vector Classification  (SVC)\footnote{sklearn.svm.SVC function, scikit-learn 0.20.4}.
        As setting of SVM, the hyperparameters $C$ and $\gamma$ are determined by grit search, and the results are adopted when the classification rate is the highest.
        Also, one-versus-the-rest classifier is used as a multiclassifier, and the classification rate is evaluated by stratified 10-fold cross-validation.
        We also used a t-test to determine whether there is a statistically significant difference in the classification rate due to vibrations.

    \subsection{Results}
    
        \subsubsection{Grit size classification}
        Figure \ref{fig: result_sandpaper} shows the experimental results for classification.
        In Figs. \ref{fig: result_sandpaper}(a) and (b), the horizontal axis is the vibration level, and the vertical axis is the class separation and SVM classification rate, respectively.
        The class separation and classification rates rise with the vibration level.
        The trends of the two measures are almost identical, with maximum values around vibration levels 4 and 5.
        The classification rate improved to about 70$\mathrm{\%}$.
        
\begin{table*}[tbh]
 \centering
 \caption{Average contact force and standard deviation in experiments}
 \label{tbl: pdc}
 \begin{tabular}{|c||c|c|c|}
  \hline
	Subject	&\begin{tabular}{c}Pressure [kPa]\\(Vibration level: 0)\end{tabular} &\begin{tabular}{c}Pressure [kPa]\\(Vibration level: 4)\end{tabular} &\begin{tabular}{c}Pressure [kPa]\\(Sliding motion)\end{tabular}\\\hline\hline
	Sandpaper  	&10.06$\pm$0.77 &10.05$\pm$0.77 &9.01$\pm$0.87\\\hline
	Slit        &14.36$\pm$1.51 &14.35$\pm$1.52 &14.79$\pm$1.51\\
  \hline
 \end{tabular}
\end{table*}
        
        As shown in Fig. \ref{fig: result_sandpaper}(c), the classification rates of the proposed method are comparable to those for the sliding motion.
        The classification rates for both cases are significantly higher than those for vibration level 0 (p $<$ 0.01).
        The confusion matrix for each case is shown in Figs. \ref{fig: result_sandpaper}(d), (e), and (f).
        In the sliding motion, misclassification occurs in the neighboring class.
        On the other hand, in the proposed method, misclassification occurs in the widely spanned classes.
        
        Figure \ref{fig: wave_sandpaper} shows the sensor output waveform and its frequency spectrum at grit 40.
        Even when no vibration is applied, a small sinusoidal wave with a small amplitude is detected may be due to the vibration caused by the robot's joint motors. 
        
        Figure \ref{fig: fspace_sandpaper} shows the feature space after dimensionality reduction by Fisher's linear discriminant.
        Fig. \ref{fig: fspace_sandpaper}(a) shows vibration level 4 and (b) shows vibration level 0. It can be confirmed that injected vibration with level 4 makes the separation among classes clearer.

        \subsubsection{Gap width classification}
        Figure \ref{fig: result_slit} shows the experimental results for classification.
        In Figs. \ref{fig: result_slit}(a) and (b), the horizontal axis is the vibration level, and the vertical axis is the class separation and SVM classification rate, respectively.
        The class separation and classification rates increase with the vibration level.
        The trends of the two measures are almost identical, with maximum values around vibration levels 4 and 5.
        The classification rate improved to about 99$\mathrm{\%}$.
        
        As shown in Fig. \ref{fig: result_slit}(c), the classification rates of the proposed method are better than those for the sliding motion.
        The classification rates for both cases are significantly higher than those for vibration level 0 (p $<$ 0.01). Moreover, the rate with the vibration level 4 is significantly higher than that for the sliding motion. 
        The confusion matrix for each case is shown in Figs. \ref{fig: result_slit}(d), (e), and (f). 
        In the sliding motion, misclassification occurs in the neighboring class.
        
        Figure \ref{fig: wave_slit} shows the sensor output waveform and its frequency spectrum at a 0.5-mm gap width.
        Even when no vibration is applied, a small sinusoidal wave with a small amplitude is detected similarly to those in the grit size classification. 
        
        Figure \ref{fig: fspace_slit} shows the feature space after dimensionality reduction by Fisher's linear discriminant.
        Figs. \ref{fig: fspace_slit}(a) and (b) show vibration levels 4 and 0. It can be more obviously confirmed than the case of the grit size classification that injected vibration with level 4 makes the separation among classes clearer.

        In summary, our experimental results using the prototype system confirm that the proposed method accomplished grit size and gap width classification tasks with comparable or better accuracy than those using a sliding motion (Figs. \ref{fig: result_sandpaper}(a), (b), and (c) and Figs. \ref{fig: result_slit}(a), (b), and (c)).
        We also confirmed that the frequency spectrum of propagated-vibration data was transformed into classifiable feature space by applying vibration (Figs. \ref{fig: fspace_sandpaper} and \ref{fig: fspace_slit}).
        Table \ref{tbl: pdc} shows that the contact force during the experiment did not change significantly depending on the experimental conditions.

\begin{figure}[tbp]
  \centering
  \includegraphics[width=\hsize]{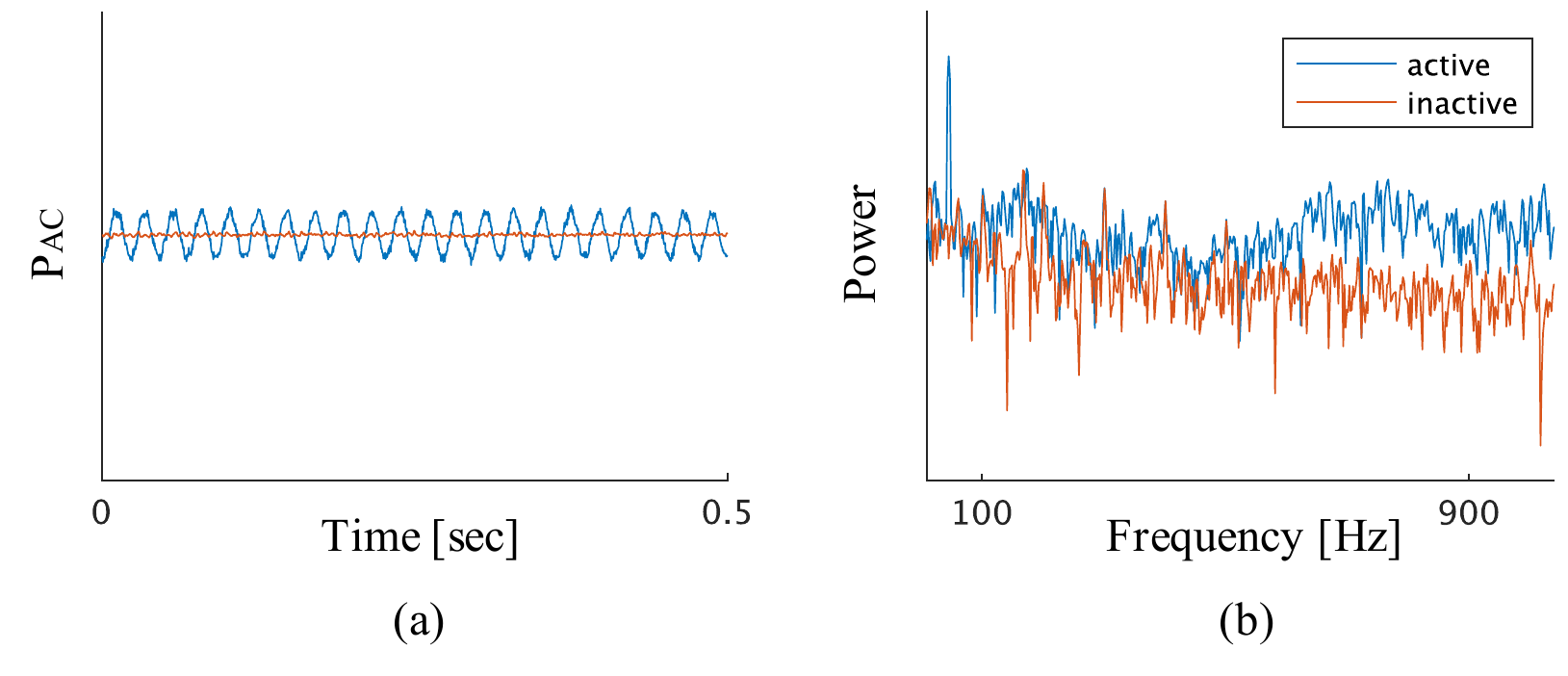}
  \caption{Output data from sensor without vibration injection: (a) micro-vibration waveform and (b) micro-vibration frequency spectrum}
  \label{fig: URwave}
\end{figure}

\section{DISCUSSIONS}
The proposed method's classification rate achieved higher accuracy with gap width classification than with grit size classification.
On the other hand, the maximum class separation and classification accuracy are maximized around vibration level 4.
The reason for this may be that at vibration levels 6 and 7, the classification task is relatively poorly accomplished because it is buried by white noise more than the propagating vibration that contains tactile information.

The misclassification in the proposed method was sparse (Figs. \ref{fig: result_sandpaper}(e) and \ref{fig: result_slit}(e)).
These results might have been caused by the table's anisotropic characteristics due to its fabrication by a 3D printer.
As the vibrations propagate to the surface texture and the table material, the classes placed across from the circular table would be expected to contain similar propagating properties. 
A method for filtering such non-tactile information needs to be investigated in future work. 
Furthermore, at vibration level 0, when no vibration was applied, we theoretically expected a classification rate of 20$\mathrm{\%}$; but it actually reached around 40$\mathrm{\%}$ (Figs. \ref{fig: result_sandpaper}(c) and \ref{fig: result_slit}(c)).
This result was probably caused by the vibration of the joint motor spreads and becomes a kind of applied vibration (Figs. \ref{fig: wave_sandpaper} and \ref{fig: wave_slit}).
We conducted an additional experiment to identify the source of the vibration.
We put the BioTac in contact with the table without applying any vibration, and investigated how the vibration waveform changed when the robot was turned on and off. 
Figure \ref{fig: URwave} shows the vibration waveform.
When the power is turned on, periodic low-frequency vibration is observed, but when it is turned off, the spectrum in the low-frequency range disappears.
This indicates that the cause of the vibration is the robot arm in the activated state.
On the other hand, a compliant attachment mechanism to the robot arm may mitigate the vibration.

It may be more desirable to attach the piezo actuator in a position that occupies as little workspace as possible. 
To this end, it is necessary to confirm whether the proposed method is effective even from a place far from the contact area, such as a nail.

The proposed method demonstrates potential for extracting tactile information in environments where the sliding motion is restricted.
It is also expected to make it possible to achieve complex tasks in environments where tactile control was impossible before.
We believe that our proposed method can be an effective tactile sensing method for tasks in narrow spaces, such as surgical procedures and working in pipes.
Besides, for such practical robot tasks, it may be necessary to exploit more complex discriminators such as deep neural networks rather than the SVM. 

In addition to grit size and gap width, we could expand the application field if we can show the effectiveness of the classification of such texture information as stickiness and slipperiness.
We also need to study how the classification task's accuracy changes when we apply it to other tactile sensors than BioTac.
As an applied vibration, we could use other signals instead of white noise, such as pink noise.
It can be considered as a direction of our future work to use the contact force as an experimental parameter by force control as well as the vibration level.

\section{CONCLUSION}
In this paper, we proposed a vibration-based tactile sensing system that does not require a sliding motion.
In the proposed method, we measured the tactile information of a subject by applying vibration to a tactile sensor's soft structure.
Our prototype system uses a biomimetic tactile sensor, a piezoelectric actuator, and the frequency spectrum obtained from the sensor as a feature.
In verification experiments, we performed two classification tasks on grit size and gap width and confirmed that the proposed method accomplished better classification rates than sliding motion.
These results show that the proposed method captured tactile information in an environment where sliding motion is infeasible.

\addtolength{\textheight}{-2cm}   

\bibliographystyle{IEEEtran}
\bibliography{IEEEabrv,ref}

\end{document}